\documentclass[conference]{IEEEtran}
\IEEEoverridecommandlockouts

\def\bng{\bngx}

\font\bngx=bang10

\def\*#1*#2{o\null{#2}{#1}}

\def\sh#1{\setbox0=\hbox{#1}%
     \kern-.02em\copy0\kern-\wd0
     \kern.04em\copy0\kern-\wd0
     \kern-.02em\raise.0433em\box0 }

\usepackage{url}
\usepackage{array}
\usepackage{cite}
\usepackage{fancyhdr} 
\usepackage{amsmath,amssymb,amsfonts}
\usepackage{algorithmic}
\usepackage{graphicx}
\usepackage{textcomp}
\usepackage{xcolor}
\def\BibTeX{{\rm B\kern-.05em{\sc i\kern-.025em b}\kern-.08em
    T\kern-.1667em\lower.7ex\hbox{E}\kern-.125em}
}

\begin{document}

\title{Sentiment Polarity Analysis of Bangla Food Reviews Using Machine and Deep Learning Algorithms
}
\author{
\IEEEauthorblockN{Al Amin\textsuperscript{1},  Anik Sarkar\textsuperscript{2}, Md Mahamodul Islam\textsuperscript{3}, Asif Ahammad Miazee\textsuperscript{4}, Md Robiul Islam\textsuperscript{5}, Md Mahmudul Hoque\textsuperscript{6}}

\IEEEauthorblockA{{\textsuperscript{1,5} Department of Computer Science and Engineering, Uttara University, Dhaka, Bangladesh} \\
{\textsuperscript{2} National Institute of Technology, Silcar, India} \\
{\textsuperscript{3} Department of Informatics, University of Oslo, Oslo, Norway} \\
{\textsuperscript{4} Department of Computer Science, Maharishi International University, Iowa, USA}\\
{\textsuperscript{6} Department of Computer Science and Engineering, CCN University of Science \& Technology, Comilla, Bangladesh}\\
\{alaminbhuyan321,  aniksarkar.cs, mdmahamodul1998, asifahammad7,robiul.cse.uu , cse.mahmud.evan \}@gmail.com}

}
\maketitle

\begin{abstract}
The Internet has become an essential tool for people in the modern world. Humans, like all living organisms, have essential requirements for survival. These include access to atmospheric oxygen, potable water, protective shelter, and sustenance.  The constant flux of the world is making our existence less complicated. A significant portion of the population utilizes online food ordering services to have meals delivered to their residences. Although there are numerous methods for ordering food, customers sometimes experience disappointment with the food they receive. Our endeavor was to establish a model that could determine if food is of good or poor quality. We compiled an extensive dataset of over 1484 online reviews from prominent food ordering platforms, including Food Panda and HungryNaki. Leveraging the collected data, a rigorous assessment of various deep learning and machine learning techniques was performed to determine the most accurate approach for predicting food quality. Out of all the algorithms evaluated, logistic regression emerged as the most accurate, achieving an impressive 90.91\% accuracy. The review offers valuable insights that will guide the user in deciding whether or not to order the food.
\end{abstract}


\section{Introduction}

Text-based opinion analysis is a branch of NLP that uses computational methods to analyze subjective information in text. This method, applied to unstructured text data, enables organizations to extract meaningful and socio-emotional information. As we approach the Internet era and the next phase of technological advancement, it is evident that text-based reviews
shared across different online platforms, particularly restaurants via online meal delivery platforms, are prevalent. Text-based reviews found online express what consumers think about a specific product. A user can easily share his or her thoughts by reviewing the food. Nowadays, people are so smart that when they are ready to order any food, they first check the reviews of the food. If most of the users gave positive feedback, then users can easily be sure that the food is good in taste. So, it is very crucial for businessmen or restaurant owners to make sure the food is of good quality. People are now less conscious of money and more concerned about the healthiness of food. If the food is good, then they are ready to pay a large amount of money to buy it. It’s essential for the owners to provide good food to the user so that they will be satisfied and give a good review because the next customer will depend on the previous customer as the internet usability is growing so rapidly and almost everyone has access to the internet. The market position of a particular restaurant can be ascertained in order to gauge user opinion. New user reviews will be categorized by the model according to their target level and the total number of reviews. positive or negative \cite{one}. Nowadays, the majority of restaurants have pages on Facebook or other review websites where patrons can leave comments. Once again, accurate review analysis is essential to a business's success. If the restaurant's management is unable to resolve the issue raised by customers in reviews, it will be difficult for the authority to remedy the problem.

According to the source \cite{two}, the assessed number of web clients in Bangladesh by January 2023 is 66.94 million. They generally write reviews in English, Bangla, or a combination of the two, with the latter being more common among smartphone users. There are no clear guidelines for posting reviews, so it is impossible to carefully review them, and there are frequently a ton of remarks made. Therefore, it may be useful to have A tool that can measure the emotional valence of text. Both positive and negative aspects of the user's affective state can be found in Bangla. Over 1484 reviews were collected for our study from numerous online food delivery websites, such as Food Panda. We strive to design a machine learning architecture that employs sentiment analysis and natural language processing (NLP) to automatically detect and display the ratio of negative to positive reviews.  We preprocessed the data by eliminating stop words, tokenization, stemming, superfluous punctuation, and other symbols. We employ CountVectorizer, Term Frequency-Inverse Document Frequency vectorizer, and N-gram to vectorize text into feature vectors, which is a means of evaluating deep learning and machine learning techniques.  Finally, we preferred the Random Forest Classifier, Linear SVM, Naïve Bayes, Decision Tree Classifier, Logistic Regression, Multinomial Naïve Bayes, and LSTM for classifying positive and negative evaluations. In conclusion, we collected over 1484 reviews of Bengali food for our study without converting or altering the original content, which the public will have access to for upcoming studies. These evaluations were gathered by hand from online meal delivery services. In order to achieve greater accuracy than just machine learning techniques, we employed deep learning techniques.
Overall, this research project's contribution is-
\begin{itemize}
    \item We gathered diverse Bengali data from several well-known online food delivery services in order to apply sentiment analysis of food reviews in Bengali text
    \item The gathered dataset was carefully annotated with two sentiment polarities: positive and negative.
    \item This much data on online Bangla food reviews in Bengali text has never been used before.
    \item Our suggested strategy outperforms existing approaches employed by researchers in terms of accuracy.
\end{itemize}

The rest of the work is separated into four segments. Segment II keeps works that are correlated. Our recommended methodology is presented in Segment III. Segment IV displays the findings of the experiment. Lastly, in Segment V, we came to a conclusion and suggested some next steps.

\section{RELATED WORKS}
In this age of advanced technology, online food ordering systems are very common. It’s becoming popular with all kinds of customers because of the easy ordering process and because people can easily pay the bill through an online bank. Sentiment analysis, also referred to as opinion mining, is a natural language processing (NLP) method employed to ascertain the emotional tone or sentiment expressed in a piece of text. Sentiment analysis seeks to comprehend and categorize the sentiment or opinion conveyed by the text as positive or negative. It can provide insights into the overall sentiment of a document, sentence, or even individual words. Several researchers have worked on this not only in food but also in other domains such as restaurants, books, movies, etc. Junaid et al. \cite{three} proposed an LSTM model after trying several machine learning and deep learning algorithms, where their best accuracy was 90.89\% with word2sequence feature extraction. Hossain et al. \cite{four} proposed a model after collecting 1,000 restaurant reviews from the Priyao website. They’ve used POs tagged and TF-IDF vectorizers for feature extraction. After using several machine learning algorithms such as multinomial Naive Bayes, logistic regression, K-Nearest Neighbor, and support vector machines, 77\% accuracy was obtained in the validation dataset by Asiful et al. \cite{five}. The Facebook page Food Bank was used for collecting the dataset. Hasan et al. \cite{six} used several machine learning algorithms on 5000 restaurant reviews, where several feature extraction methods were used, like bag of words, term frequency-inverse term frequency (TF-IDF), and skim-gram. Skip-gram gives the best accuracy among the. Following the application of LSTM and the CNN attention mechanism to the English dataset, Bhuiyan et al. achieved 98.4\% accuracy \cite{seven}. The majority of these interviews were utilized for Bangla review sentiment analysis. After collecting data from several translations or groups and translating from English to Bangla with two classes, such as positive and negative Sharif et al. \cite{eight} account for the Naive Bayes in the validation set.  After translating online shopping reviews, Rahman et al. \cite{nine} achieved 83\% accuracy for the CNN model and 78\% correctness for the Support Vector Machine (SVM). Nayan Banik et al. \cite{ten} developed a polarity detection system for movie reviews. They used a precision of 0.86 after using the stemmed unigram feature extraction technique with support vector machines and Naive Bayes. Another work has been done with restaurant reviews. Fabliha et al. \cite{eleven} developed a model with various feature extraction approaches such as N-gram techniques, count vectorizers, and TF-IDF. After using several machine learning algorithms, they got 75.58\% accuracy on SVM. Sentiment analysis not only relies on food or restaurants; it is also used in the Bangla Cricket Commentary. Mahtab et al. \cite{twelve} proposed a model where they used an SVM classifier and a TF-IDF vectorizer. They couldn’t achieve significant accuracy. Opinion analysis used not only Bangla or English but also the native language as well. In their \cite{thirteen} work, Rohini et al. present a model for identifying emotion in Kannada, an Indian native language. Machine learning algorithms are applied for classification; they show the accuracy comparison between English and Kannada datasets. Despite their inability to collaborate with other Indian languages, Animesh and Pintu \cite{fourteen} suggested a model in which they used Mutual Information to carry out feature selection \cite{fifteen}. They utilized MNB for classification and demonstrated that the Bangla dataset is more accurate than the English dataset. Nonetheless, they were unable to process actual Bangla reviews. Another work has been done with Twitter posts; they perform their work on a specific application called Traveloka. After collecting 12 thousand tweets and performing numerous machine learning algorithms, SVM got the highest accuracy \cite{sixteen}. The TripAdvisor website was used for collecting reviews of the restaurants in Manoda City. They’ve collected overall 640 reviews from that website, and they got 76.20\% accuracy for Naive Bayes \cite{seventeen}.

\section{METHODOLOGY}
\begin{figure}[ht]
    \centering
    \includegraphics[scale=0.4]{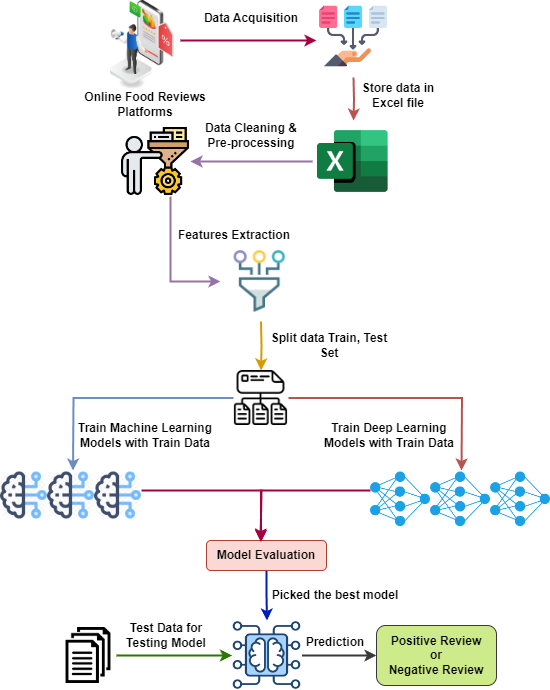}
    \caption{Proposed Methodology Workflows.}
    \label{fig:pmethodology}
\end{figure}

Within this section, an exhaustive account of the comprehensive approach and the suite of tools employed for the execution of the sentiment analysis task on Bengali Foods reviews is delineated. Before commencing the training process, it is imperative to elaborate on the dataset pre-processing steps. Subsequently, we will introduce our proposed model for sentiment analysis. The workflow, as elucidated in Figure \ref{fig:pmethodology}, provides a visual representation of the entire process, with further intricacies expounded upon in the subsequent sections.

\begin{subsection}{Data Acquisition}
Convocating, calculating, and checking exact observations for research based on recognized systems constitutes the process of data gathering. Researchers can derive their hypotheses from the information they have collected. The importance of data collection has a direct bearing on the upkeep of research unity and the making of business judgments. Our data was gathered from online reviews found on several online meal delivery services, including FoodPanda, Shohoz Food, Pathao Food, HungryNaki, and others. We collected this data manually from different restaurant and food review sections under the online food delivery services without changing or violating the original review. Then we stored those data in an Excel file and their corresponding sentiments. A total of 1400+ Bengali food reviews have been gathered, of which 718 have been classified as positive and 766 as negative. Later, our dataset contains two columns’ reviews and sentiment. The task is a binary classification problem to detect whether the review is classified as positive 1 or negative 0. We can see some examples of data in Table \ref{table:dataExample}, and Table \ref{table:dataCount} shows the distribution for both the positive and negative classes of our dataset.
\begin{table}[ht]
\caption{Example of Data}\label{table:dataExample}
\centering
\begin{tabular}{|c|p{5cm}|c|}
\hline
\textbf{NO.} & \textbf{Reviews} & \textbf{Sentiment} \\ [1ex] \hline
1 & {\bng Ek baer baej khabar Et baej khabar Aaim bhabet{O} pair na{I}.}  & 0 \\ \hline
2 & {\bng {O}Jel phueDr khabar mjar ichl EbNNG AacrN {O} bhal ichl.} & 1 \\ \hline
\end{tabular}
\end{table}

\begin{table}[ht]
\caption{Distribution of Classes}\label{table:dataCount}
\centering
\begin{tabular}{|c|p{3cm}|}
\hline
\textbf{Review Sentiment Polarity} & \textbf{Number of Data} \\ [1ex]
\hline
Positive & 718 \\
\hline
Negative & 766  \\
\hline
\textbf{Total} & \textbf{1484}  \\
\hline
\end{tabular}
\end{table}
\end{subsection}

\begin{subsection}{Data Cleaning and Pre-processing}
    Data wrangling is the process of converting raw data into a usable format for machine learning algorithms. Cleaning the data is far too crucial in order to extract high-quality data for our experiment, which will enable us to minimize test errors. Textual representations of online food reviews are not sufficient for machine learning. We must make sure that the data has been cleansed before we train the model. Our dataset has to be pre-processed using the Bangla text pre-processing approach in order to get the textual representations ready for model building. First, include contractions in our dataset, and then extract the contractions' brief form. Using regular expressions, we first eliminated punctuation marks, emoticons, pictorial icons, emojis, random English words, and alphabets from our dataset. There are also various Unicode characters in Bangla text that we must precisely put back. Second, we employ the BNLP Python library to tokenize our assessment transform phrases into individual words and retrieve Bangla language-specific exclusionary terms. Furthermore, we extract the root words from each word token using Bangla word stemming methods. Finally, all related words and, to improve feature extraction, word stems are amalgamated into sentences. Below Table \ref{table:preprocessData} is a demonstration of raw data and cleaned data differences.
    \begin{table}[ht]
    \caption{Comparison of Raw Reviews and Cleaned Reviews}
    \label{table:preprocessData}
    \centering
    \begin{tabular}{|c|p{4cm}|p{3cm}|}
    \hline
    \textbf{NO.} & \textbf{Raw Reviews} & \textbf{Cleaned Reviews} \\ [1ex] \hline
    1 & {\bng {O}eyTar edr bYbHar Ekdm phaltu,,,, {I}Ta taedr kaes Aasa isl na,,, pura{I} jghnY bYbHar...... raibsh} & {\bng {O}eyTar edr bYbHar Ekdm phaltu {I}Ta taedr kaes Aasa isl na pura{I} jghnY bYbHar raibsh} \\ \hline
    2 & {\bng Airijnal baribikU icekn  EbNNG Hain mas/Tar/D ss !!! kha{O}yar jnY ineyichlam sWadTa cmt//kar ichl EbNNG pirebsh{O} khub bhal ichl} & {\bng Airijnal  baribikU icekn EbNNG Hain mas/Tar/D ss kha{O}yar jnY ineyichlam sWadTa cmt//kar ichl EbNNG pirebsh{O} khub bhal ichl} \\ \hline
    \end{tabular}
    \end{table}
\end{subsection}
\begin{subsection}{Features Distillation}
    To feed the classifier models, several features were taken out of the review text. As previously indicated, we classified our meal review dataset using both deep learning models and conventional machine learning techniques. Textual representations cannot be classified by either of these models. They require numerical representations of numerous textual and sentence-level qualities and attributes to analyze and forecast distinct feelings. To explore and obtain better results, we employed different pieces for different classifiers. In this part, we'll quickly review those features. \\
    
    \textbf{Counter Vectorizer:} Count Vectorizer facilitates the conversion of text data into a matrix format, where each row represents a document and each column represents a word, with the matrix elements representing the word counts. The count vectorizer creates a sparse matrix of all the unique words in the dataset. In a document, every word has an index number that indicates how often that word appears in that document. Then, we use the vector delineation as a feature for diverse models. We applied count vectors to our dataset and tested them on abundant imitation. \\
    Let \textbf{\textit{D}} be the collection of documents and \textbf{\textit{t}} be a term (word). The count vectorization of term \textbf{\textit{t}} in document \textbf{\textit{d}} is denoted as \textbf{\textit{Count(t,d)}}. The matrix representation \textbf{\textit{X}} is created by counting the occurrences of terms in documents.
    \begin{equation}
    \label{eq:countVectorizer}
    X_{t,d} = \text{Count}(t,d)
    \end{equation}
    Here, $X_{t,d}$ represents the count of term \textbf{\textit{t}} in document \textbf{\textit{d}}. \\
    
     \textbf{TF-IDF:} TF-IDF stands for term frequency-inverse document frequency. It is a statistical measure that evaluates how important a word is to a document in a collection of documents. This is done by multiplying two metrics. Term frequency (TF): This is the number of times a word appears in a document. Inverse document frequency (IDF): This is a measure of how rare a word is across a set of documents. The rarer a word is, the more important it is to a document. For TF-IDF vectorizing, we employed bigram and unigram, with contiguous two words serving as additional feature input. N-grams are helpful in creating context between words, which helps to convey better sentence qualities. The \textbf{\textit{"tf-idf"}} statistic we're using is,
     \begin{equation}
     \label{eq:tfidf}
      \text{\textit{tf-idf(w, r)}} = tf(w, r) \frac{N}{|\{r \in R: w \in r\}|}
     \end{equation}
     
     Here,
     \begin{itemize}
         \item \textit{tf-idf(w,r)} = value of word w in the review \textbf{\textit{r}},
         \item \textit{tf(w,r)} = frequency of word w in review \textbf{\textit{r}},
         \item \textit{N} = total amount of reviews,
         \item \textit{$|\{r \in R: w \in r\}|$} = number of reviews containing \textbf{\textit{w}}.
     \end{itemize}

     \textbf{GloVe-Vectors:}  This is another approach to represent a word in vector space. GloVe is a technique for unsupervised learning that generates vector representations of words. The training process is carried out using combined global word-word co-occurrence statistics from a corpus. The outputs demonstrate intriguing linear substructures inside the word vector space. In our study, we use Bengali pre-trained GloVe vector that is trained on Wikipedia+crawl news articles (39M (39055685) tokens, 0.18M (178152) vocab size. This model is built for BNLP package. BNLP is a natural language processing toolkit for Bengali Language. This tool will help you to tokenize Bengali text, Embedding Bengali words, Embedding Bengali Document, Bengali POS Tagging, Bengali Name Entity Recognition, Bangla Text Cleaning for Bengali NLP purposes. This model is available on the HuggingFace platform.
\end{subsection}
\begin{subsection}{Sentiment Polarity Models}
    In our study, we split the dataset training and test set as 80\% and 20\% ratio. We applied different well-known machine learning algorithms such as Logistic-Regression (LR), DecisionTreeClassifier, RandomForestClassifier, MultinomialNB, KNeighborsClassifier, Support Vector Classification (SVC), SGD Classifier or Stochastic Gradient Descent (SGD). We also tried deep learning approach to get good results for that we used the Recurrent Neural Network (RNN), Long Short-Term Memory (LSTM), and Gated Recurrent Unit (GRU) model. We used some hyperparameters to tune the model to get better accuracy. The hyperparameter value is given in the below table for machine learning and deep learning models.
\end{subsection}
\begin{subsection}{Model Implementation}
   We've gathered various attributes such as Tf-Idf, glove vector, count vector, and more from the refined raw text information. Following this, we input the data into our deep learning and machine learning models. The machine learning model, also referred to as the logistic regression model, has produced results that are reasonably excellent. We trained our Logistic Regression model with random state 123 using the unigram features vectors. Therefore, our accuracy is 90.91\%.
\end{subsection}
\section{EXPERIMENTAL RESULTS}
In this research for the training model, data was split into 80\% for training and 20\% for testing. We have examined several machine learning and deep learning models. To assess our model's effectiveness, we've employed four distinct categories of metrics. These metrics are outlined below. \\

\textbf{Accuracy:} is one crucial factor in evaluating the classifier's performance. This metric indicates the proportion of correct classifications among all predictions made by the classifier. Accuracy calculation is defined in Equation-\ref{eq:1}.
\begin{equation}
\label{eq:1}
\text{Accuracy} = \frac{TP + TN}{TP + TN + FP + FN}
\end{equation} \\

\textbf{Precision:} can be described as the count of accurate positive outputs generated by the model. The precision calculation is shown in Equation-\ref{eq:2}.
\begin{equation}
\label{eq:2}
\text{Precision} = \frac{TP}{TP + FP}
\end{equation} \\

\textbf{Recall:} measures the effectiveness of a model in recognizing positive instances among all the genuine positive instances within the dataset. Recall calculation is defined in Equation-\ref{eq:3}.
\begin{equation}
\label{eq:3}
\text{Recall} = \frac{TP}{TP + FN}
\end{equation} \\

\begin{figure}[ht]
    \centering
    \includegraphics[scale=0.77]{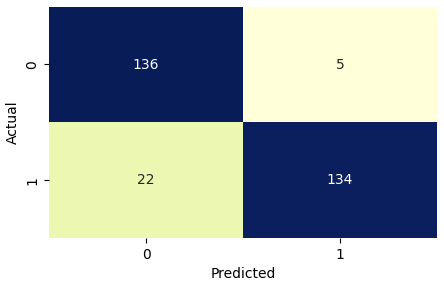}
    \caption{Confusion Matrix of Proposed Model}
    \label{fig:CM}
\end{figure}

\begin{figure}[ht]
    \centering
    \includegraphics[scale=0.85]{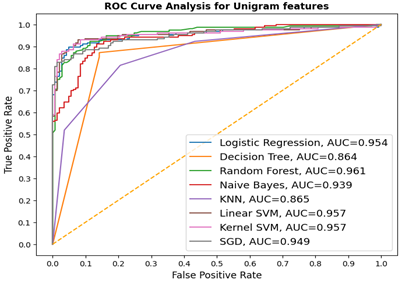}
    \caption{ROC Curve of Proposed Model}
    \label{fig:ROCCurve}
\end{figure}

\textbf{F1-Score:} assists in assessing both recall and precision concurrently. The F1-score reaches its peak when recall matches precision. It can be calculated using Equation-\ref{eq:4}.
\begin{equation}
\label{eq:4}
\text{F1-Score} = \frac{2 * Precision * Recall}{Precision + Recall}
\end{equation} \\
The subsequent parameters are employed in computing the aforementioned metrics:
\begin{itemize}
    \item \textbf{True Positive(TP):} The model has predicted a sentiment as positive, and the actual sentiment was positive. 
    \item \textbf{False Positive (FP):} The model has predicted a sentiment positive, but the actual sentiment was negative.
    \item \textbf{True Negative (TN):} The model has predicted a sentiment negative, and the actual sentiment was also negative. 
    \item \textbf{False Negative (FN):} The model has predicted a sentiment was negative and actually sentiment was positive. 
\end{itemize}
\begin{subsection}{Machine Learning Model Evaluation}
    Table-\ref{table:MLModelAcc} displays the machine learning model's output. Tf-Idf with unigram was utilised in this experiment as a feature extraction method to feed the machine learning models. It is evident that Logistic Regression performs fairly well when a Tf-Idf is employed.
    \begin{table}[ht]
    \caption{Research comparison with others}\label{table:modelCompare}
    \centering
    \begin{tabular}{|c|p{1cm}|c|c|c|c|}
    \hline
    \textbf{No.} & \textbf{Authors} & \textbf{Topic} & \textbf{Data} & \textbf{Model} & \textbf{Accuracy} \\ [1ex]
    \hline
    1 & Junaid et al . [2]  & Food & 1040 & LSTM & 90.86\% \\ [1ex]
    \hline
    2 & O. Sharif et al [7]  & Restaurent & 1000 & MNB & 80.48\% \\ [1ex]
    \hline
    3 & \textbf{Proposed Model} & \textbf{Food} & \textbf{1484} & \textbf{LR} & \textbf{90.91\%} \\ [1ex]
    \hline
    \end{tabular}
    \end{table}
    \begin{table}[ht]
    \caption{Model Evaluation table}\label{table:MLModelAcc}
    \centering
    \begin{tabular}{|c|p{2cm}|c|c|c|c|}
    \hline
    \textbf{No.} & \textbf{Algorithm} & \textbf{Accuracy} & \textbf{Precision} & \textbf{Rcall} & \textbf{F1-Score} \\ [1.5ex]
    \hline
    1 & Logistic Regression & 90.91 & 96.40 & 85.90 & 90.85 \\ [1.5ex]
    \hline
    2 & DecisionTree Classifier & 85.52 & 86.93 & 85.26 & 86.08 \\ [1.5ex]
    \hline
    3 & RandomForest Classifier & 89.56 & 94.33 & 85.26 & 89.56 \\ [1.5ex]
    \hline
    4 & MultinomialNB & 84.51 & 91.04 & 78.21 & 84.14 \\ [1.5ex]
    \hline
    5 & KNeighbors Classifier & 80.47 & 81.41 & 81.41 & 81.41 \\ [1.5ex]
    \hline
    6 & Linear SupportVectorClassification & 84.90 & 98.46 & 82.05 & 89.51 \\ [1.5ex]
    \hline
    7 & RBF SupportVectorClassification & 89.90 & 98.46 & 82.05 & 89.51 \\ [1.5ex]
    \hline
    8 & SGD Classifier & 90.24 & 95.04 & 85.90 & 90.24 \\ [1.5ex]
    \hline
    \end{tabular}
    \end{table}
\end{subsection}
\\
Our model's confusion matrix is displayed in Fig-\ref{fig:CM}. Here, 0 and 1 stand for negative and positive values, respectively. In this figure, we can clearly see the false positive and false negative values are very low in comparison to true negative and true positive. This indicates that our model is more accurate.
 \\
Fig-\ref{fig:ROCCurve} showing the ROC curve analysis for unigram features in different machine learning models.

\begin{figure}[ht]
    \centering
    \includegraphics[scale=0.85]{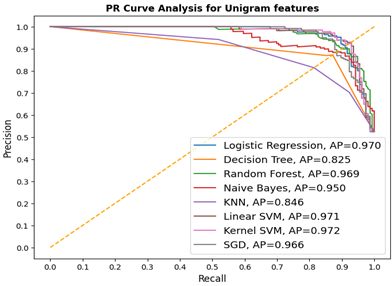}
    \caption{Precision, Recall Curve of Proposed Model}
    \label{fig:PRCurve}
\end{figure}

Fig-\ref{fig:PRCurve} showing the Precision and Recall curve analysis for unigram features in different machine learning models.
\section{CONCLUSION AND FUTURE WORK}
Our research aimed to examine public sentiment towards Bengali cuisine through the analysis of online reviews. As internet usage has risen in Bangladesh, individuals have increasingly turned to online platforms to express their opinions and experiences regarding food. We collected 1400+ Bengali (Bengali Text) food reviews from social media and other online platforms and trained them using a range of machine learning and deep learning techniques. To make the text reviews easier for comprehension by our algorithm, we used some data cleaning and feature extraction techniques before the model training. A comparative evaluation of all the classifiers is carried out using feature extraction methods like TF-IDF. According to the results of our experiment, the Logistic Regression (LR) classifier can reach the greatest accuracy of TF-IDF, at 90.91\%.
\newline
Even though we received a decent score for the proposed model, expanding the dataset will yield better results. In the future, mobile or web-based applications can be developed using pre-trained models based on BERT or Transformer to obtain greater performance. In this study, we worked on binary class classifications, positive and negative. So, in future work, it can be expanded to include multi-class classification problems like positive, negative, and neutral sentiment polarity analysis. Since our dataset contains only positive and negative classes. So, by adding some neutral reviews, it can be possible to solve a multi-class classification problem.

\bibliographystyle{ieeetr}
\bibliography{references}
\end{document}